\documentclass{article}

\usepackage{PRIMEarxiv}

\usepackage[utf8]{inputenc} 
\usepackage[T1]{fontenc}    
\usepackage{hyperref}       
\usepackage{url}            
\usepackage{booktabs}       
\usepackage{amsfonts}       
\usepackage{nicefrac}       
\usepackage{microtype}      
\usepackage{lipsum}
\usepackage{fancyhdr}       
\usepackage{graphicx}       
\graphicspath{{media/}}     
\usepackage{placeins}

\title{Exploring Linguistic Similarity and Zero-Shot Learning for Multilingual Translation of Dravidian Languages}

\author{
 Danish Ebadulla \\
  PES University\\
  \texttt{danish.ebadulla@gmail.com} \\
   \And
 Rahul Raman \\
  PES University\\
  \texttt{rahulraman1604@gmail.com} \\
  \And
 S. Natarajan \\
  PES University\\
  \texttt{natarajan@pes.edu} \\
  \And
 Hridhay Kiran Shetty \\
  PES University\\
  \texttt{hridhayks@gmail.com} \\
   \And
 Ashish Harish Shenoy \\
  PES University\\
  \texttt{ashish1shenoy@gmail.com} \\
}

\begin{document}
\maketitle

\begin{abstract}
Current research in zero-shot translation is plagued by several issues such as high compute requirements, increased training time
and off target translations. Proposed remedies often come at the cost of additional data or compute requirements. Pivot based neural
machine translation is preferred over a single-encoder model for most settings despite the increased training and evaluation time. In
this work, we overcome the shortcomings of zero-shot translation by taking advantage of transliteration and linguistic similarity.
We build a single encoder-decoder neural machine translation system for Dravidian-Dravidian multilingual translation and perform
zero-shot translation. We compare the data vs zero-shot accuracy tradeoff and evaluate the performance of our vanilla method against
the current state of the art pivot based method. We also test the theory that morphologically rich languages require large vocabularies
by restricting the vocabulary using an optimal transport based technique. Our model manages to achieves scores within 3 BLEU of
large-scale pivot-based models when it is trained on 50\% of the language directions.
\end{abstract}

\section{Introduction}
India has over 121 unofficial languages in India and 22 officially recognized languages. A majority of these languages originated from 2 language families, namely Indo-Aryan and Dravidian. Over time, these languages have evolved by adopting unique characteristics such as pronunciations and dialects. Each evolution adds to the ever increasing language count of a diverse country like India. Since languages spoken in a region can be traced back to a single source, it is reasonable to assume that despite the variety in script and dialect, these languages retain certain traits that are common to their source. By exploiting this similarity, we can train deep learning models to represent them accurately in an efficient embedding space, and use their shared characteristics to allow them to support each other in natural language understanding tasks.\\
Neural Machine Translation has dominated the field of machine translation due to its superior modelling capacity. But this technique works poorly when there are very little available resources to train your model on. Low resource language research has shown that training multilingual models and enabling them to share their embeddings can help mitigate these shortcomings. But a lack of data still affects the model's ability to generalize to out-of-domain data. Indic languages are a prime example of this issue. Most Indic corpora are either poor quality or domain specific and despite the high BLEU scores that they might produce when exposed to in-domain test sets, they fail to generalize in real world scenarios. The inherent morphological richness of Indian languages only makes this problem more severe. One solution to overcome the lack of parallel data availability is to take advantage of zero-shot translation, where an NMT model is trained to translate sentences for language pairs that it has never seen before. Subword segmentation can be used to overcome the rich morphology of Indic languages by breaking up complex words into their root forms. We can also use transliteration to overcome the difference in scripts in Indic languages, thus emphasizing the phonemic similarities between languages that originate from the same family.\\

In this paper, we present a multilingual language model to translate between languages belonging to the Dravidian family. Our major contributions in this work are:
\begin{itemize}
    \item A pipeline to build a single encoder-decoder multilingual translation model for Dravidian languages that can perform zero-shot learning without implicit or explicit bridging.
    \item A study on the tradeoff between data and accuracy for zero-shot translation accuracy
    \item An evaluation of the effectiveness of transliteration and vocabulary optimization on zero-shot translation.
\end{itemize}

\section{Background}
Neural Machine Translation \cite{DBLP:journals/corr/BahdanauCB14,DBLP:journals/corr/WuSCLNMKCGMKSJL16,DBLP:journals/corr/SutskeverVL14} is currently the favored approach for machine translation tasks, as its data driven nature provides better modelling capacity, adaptability and overall performance. The data driven nature of NMT models is a double edged sword as it cannot represent language pairs that it has not seen explicitly during training with a high accuracy. Zero-shot learning \cite{firat-etal-2016-multi} is the technique used to teach NMT models to adapt to unseen language pairs. There are multiple approaches to teach your model to zero-shot translate. Google \cite{DBLP:journals/corr/JohnsonSLKWCTVW16} introduced their multilingual NMT system that can perform zero-shot translation using implicit bridging. Previous works had attempted several unique techniques to solve the problem of zero-shot learning and had been met with mixed results. \cite{dong-etal-2015-multi} proposed using a separate decoder and attention mechanism for each target language. \cite{zoph-knight-2016-multi} used a separate encoder and attention mechanism for each source language. \cite{firat-etal-2016-multi} proposed multiple encoders and decoders for each source and target language. With the increase in research on zero-shot translations, several issues that persisted came to the forefront. Off-target translations was the most tenacious issue \cite{zhang-etal-2020-improving} with the NMT models translating to a wrong target language. While \cite{DBLP:journals/corr/JohnsonSLKWCTVW16} proposed using target langauge tags that are prefixed to input sentences to alleviate this, the problem still persisted when models were trained on multiple language pairs or morphologically rich languages. Pivot based models \cite{DBLP:journals/corr/ChengLYSX16, DBLP:journals/corr/ChenLCL17,DBLP:journals/corr/abs-1811-01389} found popularity over vanilla zero-shot translation as it was possible to overcome spurious correlation \cite{DBLP:journals/corr/abs-1906-01181} and translate between languages as long as parallel corpora with a common pivot language existed. \cite{DBLP:journals/corr/abs-1903-07091,DBLP:journals/corr/abs-1805-10338} reported that vanilla zero-shot NMT is senstive to training conditions and the quality of translations are usually inferior to explicit pivot bridging. But even explicit briding has its fair share of shortcomings, as it requires double the translation time, needs 2 trained models and can potentially suffer a loss in quality if the pivot language is not linguistically similar to the source and target languages. \cite{DBLP:journals/corr/ChengLYSX16} proposed a method for jointly training both the models in a pivot-based zero-shot setting, thus cutting down the training time. Zero-shot learning for Indic languages lags far behind western languages. Indic-Indic language translation is a challenging problem as most Indian languages have large variations in their morphology and scripts, which hampers an NMT model's ability to represent the languages accurately in its embedding space. Lack of general domain corpora and standardized test sets further exacerbated this problem. Most works utilise pivot based methods \cite{DBLP:journals/corr/abs-1907-12437,10.1145/3407912}. \cite{kunchukuttan2021empirical} recently proposed a Multi-brige Multilingual NMT model for Indic-Indic translation that enables zero-shot translation. His work showed that pivot based models outperform vanilla zero-shot learning by a huge margin and that fine tuning english-centric MNMT models with a small quantity of Indic-Indic corpora produces the best results for NMT. 

\begin{table}[h]
  \centering 
  \caption{Details of the training corpora used for all experiments. All values are in 1000s of sentences}
  \label{corpora}
  \begin{tabular}{c c c c c}
    \toprule
       & kn & ml & te & ta \\
    \midrule
    kn & - & 2,852 & 2,475 & 2,776 \\
    ml & 2,852 & - & 2,650 & 2,558 \\
    te & 2,475 & 2,650 & - & 2,573 \\
    ta & 2,776 & 2,558 & 2,573 & - \\
    \bottomrule
  \end{tabular}
\end{table}

\section{Methodology}
In this work, we present multilingual Indic-Indic translation models that incorporate 4 Dravidian languages - Kannada, Tamil, Telugu and Malayalam. These 4 languages give 12 possible translation directions in an Indic-Indic language setting. We train on only a fraction of these 12 directions and evaluate on all 12 directions using zero-shot learning. The following subsections will present our pipeline in detail and highlight the unique characteristics of each component.  

\subsection{Data}
The Samanantar dataset \cite{DBLP:journals/corr/abs-2104-05596} contains parallel corpora for 11 Indic languages. Our experiments focus on Kannada, Tamil, Telugu and Malayalam, all of which are languages belonging to the Dravidian language family. Each translation direction has approximately 2.7M parallel sentences. We validate and test our models’ performance on the WAT-2021 MultiIndic Shared Task\footnote{http://lotus.kuee.kyoto-u.ac.jp/WAT/indic-multilingual/} Parallel corpora. This corpora has 11 languages and is 11-way parallel. The test set is taken from the PMI dataset and has 2360 sentences scraped from speeches given by the Prime Minister of India. Table \ref{corpora} contains the specifications of all the corpora that this work utilizes.

\subsection{Preprocessing}
All sentences are cleaned to remove foreign tokens and tokenized using the Indic NLP library. A few additional heuristics are added to account for Indic punctuation and non-breaking prefixes \cite{DBLP:journals/corr/abs-2104-05596}. We noticed that not all sentences ended with a full stop and this often affected the model’s ability to end a sentence during evaluation. [END] tags are appended to every sentence during training to help the model with identifying the end of a sentence. 

\subsection{Transliteration}
A persistent problem in zero-shot translation is off-target translations (i.e. a model translating to a wrong target language). We overcome this issue by using transliteration on our corpora. We hypothesize that languages belonging to the Dravidian family have similar phonemes but differ in their script and that transliteration in combination with subword segmentation can help reduce vocabulary size. Our experiments on this hypothesis are covered in Section \ref{overlap}. Transliteration is made possible due to the fact that Unicode points of various Indic scripts are at corresponding offsets from the base codepoint for that script. We use the IndicNLP library to transliterate our corpora and experiment with 3 different transliteration target languages. We test with a morphologically poorer script (Devanagari), the language with the smallest average sentence length in our corpora (Kannada) and the language with the largest average sentence length (Malayalam).

\subsection{Subword Segmentation}
Since all our languages are now in the same script, we learn a common vocabulary for all 4 languages. We use SentencePiece with the BPE language model and set the vocabulary size to 32K. We also experiment with VOLT \cite{xu-etal-2021-vocabulary} to learn a smaller vocabulary using optimal transport. We set the vocabulary size threshold to 10000 and let VOLT recommend the optimal vocabulary and size. A new SentencePiece model is then trained with the vocabulary size set to VOLT’s recommendation.

\subsection{Zero-shot learning}
Zero-shot translation is a technique used to translate between languages when there is little or no parallel data available for that language pair. There are 2 approaches to zero-shot translation. The most popular technique is explicit bridging; where we translate to a pivot language from the source language and then translate to the target language. The conventional choice for the pivot language is English since NMT is data driven and the largest source of parallel data for a language X is usually X-En. The other approach was proposed by \cite{DBLP:journals/corr/JohnsonSLKWCTVW16}, called implicit bridging, where a single encoder-decoder model is trained for all the language pairs and the embeddings and shared between languages in the encoder and decoder to allow for implicit ‘bridging’ which enables zero-shot translation. If there are n languages in your parallel corpora, there are n*(n-1) translation directions that are possible. These translation directions can be achieved by either explicitly training our model on parallel corpora for that pair or by exploiting zero-shot translations.\\
Our preliminary experiments showed that it is necessary for each language to be seen at least once in both the encoder and decoder during training for the model to be able to perform zero-shot translation with acceptable accuracy during evaluation.

\subsection{System Architecture}
We use the transformer model with the Transformer-Base implementation \cite{NIPS2017_3f5ee243} for multilingual translation. The models are built, trained and evaluated using the fairseq library \cite{DBLP:journals/corr/abs-1904-01038} on the translation task. We utilise the technique from \cite{DBLP:journals/corr/JohnsonSLKWCTVW16} to make use of multilingual data in a single system, by introducing a token at the beginning of the input sentence to indicate the translation direction. Since we transliterate all our language pairs into a single script, subwords with the same spelling but different meanings can be ambiguous to translate \cite{DBLP:journals/corr/JohnsonSLKWCTVW16}, hence we also add source language tags to our input sentences. The final tag is given by \_\_src\_\_x1\_\_ \_\_tgt\_\_x2\_\_ where x1 and x2 are the source and target languages respectively. To avoid out-of-vocabulary words we use subword segmentation to split the words into a vocabulary of size 32000.

\subsection{Evaluation}
We use the WAT-2021 shared task corpora for validation and testing. The dev set contains 1000 sentences and the test set contains 2390 sentences per language that are n-way parallel. 
All our models are evaluated using case-insensitive BLEU. Beam search is used with beam length set to 5. Scores are evaluated using the sacrebleu package. 

\section{Experiments and Results}
\subsection{Vocabulary Overlap}\label{overlap}
Since all our of languages belong to the Dravidian language family, we hypothesize that despite the differences in script, there should be a significant overlap in phonemes when the corpora are broken down to subwords. To test this hypothesis, we apply transliteration to all 4 languages and compare the overlap in their vocabulary. We experiment with a morphologically poorer language (Devanagari) and 2 of the source languages (Kannada and Malayalam). The subwords are generated using SentencePiece \cite{DBLP:journals/corr/abs-1808-06226} after setting the size of the vocabulary to 32000 and are then optimized to 10000 using VOLT. Table \ref{vocab-overlap} shows the percentage overlap between languages before and after transliteration. We observe that Kannada produces the highest percentage overlap for Telugu and Malayalam produces the highest overlap for Tamil. Devanagari results are largely consistent with a slight drop in overlap for Tamil. We run preliminary translation tests for all 3 scripts and document their performance in Appendix \ref{Appendix}.

\begin{table*}[h]
\caption{\label{vocab-overlap}The vocabulary overlap across language pairs. Table 1 (Top Left) represents overlap without transliteration. Table 2 (Top Right) represents vocabulary overlap with Devanagari transliteration. Table 3 (Bottom Left) represents vocabulary overlap with Kannada transliteration. Table 4 (Bottom Right) represents vocabulary overlap with Malayalam transliteration}
\centering
\begin{tabular}{c c c}
 \\
    & 
    \begin{tabular}{c|c c c c}
    \hline
    & kn & ml & te & ta \\\hline
    kn & 100.00 & 3.76 & 3.47 & 3.52 \\
    ml & - & 100.00 & 3.86 & 3.86 \\
    te & - & - & 100.00 & 2.81 \\
    ta & - & - & - & 100.00 \\
    \hline
    \end{tabular}
    &
    \begin{tabular}{c|c c c c}
    \hline
    & kn & ml & te & ta \\\hline
    kn & 100.00 & 30.64 & 31.35 & 14.85 \\
    ml & - & 100.00 & 31.30 & 19.66 \\
    te & - & - & 100.00 & 14.95 \\
    ta & - & - & - & 100.00 \\
    \hline
    \end{tabular}\\ 
    \\
  & 
    \begin{tabular}{c|c c c c}
    \hline
    & kn & ml & te & ta \\\hline
    kn & 100.00 & 22.95 & 36.00 & 12.89 \\
    ml & - & 100.00 & 28.33 & 16.78 \\
    te & - & - & 100.00 & 15.93 \\
    ta & - & - & - & 100.00 \\
    \hline
    \end{tabular}
    &
    \begin{tabular}{c|c c c c}
    \hline
    & kn & ml & te & ta \\\hline
    kn & 100.00 & 23.89 & 36.00 & 26.8 \\
    ml & - & 100.00 & 26.11 & 32.7 \\
    te & - & - & 100.00 & 24.3 \\
    ta & - & - & - & 100.00 \\
    \hline
    \end{tabular}\\ 
\end{tabular}
\end{table*}

\subsection{Training Protocol}
We train various models with increasing number of language pairs that are fed during training time to identify the optimal tradeoff between zero-shot performance and existing data requirements. The following subsections will refer to the models by the language pairs that they are trained on and the training direction. The four languages that are fed into our model are Kannada (KN), Tamil (TA), Telugu (TE) and Malayalam (ML). A double ended arrow $\leftrightarrow$ indicates that corpora was fed into the model in both directions. All our models are trained until convergence by setting patience with early stopping to 5 epochs. We run preliminary experiments to test the performance of various transliteration scripts on the translation task. Since previous literature and baselines use Devanagari, we also proceed with Devanagari as the script of choice for transliteration to obtain results comparable with baselines. We also test the role of vocabulary size in model performance by testing a multilingual model with a subword vocabulary of size 10000. This vocabulary is generated using VOLT. Vocabulary Learning via Optimal Transport (VOLT) \cite{xu-etal-2021-vocabulary} is a technique to determine the optimal vocabulary and size for a corpus without any trial training. VOLT was found to achieve upto 70\% vocabulary size reduction while still managing to achieve higher BLEU scores on various translation tasks. VOLT support vocabularies for both SentencePiece and subword-nmt \cite{DBLP:journals/corr/SennrichHB15} but we limit our tests to the former only to maintain consistency in our results.

\begin{table}[h]
\centering
  \caption{BLEU score for 4 language pair model 1. The rows are the source language and the columns are the target language. Cells in bold represent the translation directions used in training}
  \label{hi-2lang}
  \begin{tabular}{c|c c c c}
    \hline
       & kn & ml & te & ta \\
    \hline
    kn & - & \textbf{7.7} & 1.0 & 0.8 \\
    ml & \textbf{8.9} & - & 5.7 & 0.6 \\
    te & 0.5 & 3.2 & - & \textbf{7.4} \\
    ta & 5.8 & 4.9 & \textbf{7.4} & - \\
    \hline
\end{tabular}
\end{table}

\subsection{4 language pairs}
A single encoder-decoder model is trained on 4 language pairs. We run experiments with 3 types of language pair combinations:
\begin{itemize}
    \item \textbf{Model 1}: 2 unique language pairs in forward and reverse direction: kn$\leftrightarrow$ ml,te $\leftrightarrow$ ta. Results documented in Table \ref{hi-2lang}
    \item \textbf{Model 2}: 4 unique language pairs: kn$\rightarrow$ml, ml$\rightarrow$te, te$\rightarrow$ta, ta$\rightarrow$kn. Results documented in Table \ref{hi-4lang} 
    \item \textbf{Model 3}: 4 unique language pairs with VOLT: kn$\rightarrow$ml, ml$\rightarrow$te, te$\rightarrow$ta, ta$\rightarrow$kn. Results documented in Table \ref{hi-4langV}
\end{itemize}
Both techniques expose the model to 1/3 of the total translation directions during training but the first technique is built to test model performance in very low resource conditions; when there are only 2 sources of parallel corpora available. In comparison, the second model is exposed to 1/3 of the total translation directions with each source-target language combination being unique. We ensure that in every model, both the encoder and decoder see each language atleast once during training.\\
We observe that BLEU score for zero-shot translation lags by 5.03 on average compared to the performance of trained language pairs when we train on both directions of 2 language pairs only. In comparison, the zero-shot translation BLEU score lags by 5.98 BLEU on average for the model trained on 4 unique language pairs. The BLEU score for trained translation directions is always in the 6-8 BLEU range. The 4 language pair model trained with VOLT outperforms both the 32000 vocabulary models in zero-shot translation performance with zero shot scores lagging by 3.53 on average from the trained directions. 
\begin{table}[h]
\centering
  \caption{BLEU score for 4 language pair model 2. The rows are the source language and the columns are the target language. Cells in bold represent the translation directions used in training}
  \label{hi-4lang}
  \begin{tabular}{c|c c c c}
    \hline
       & kn & ml & te & ta \\
    \hline
    kn & - & \textbf{7.4} & 0.4 & 0.5 \\
    ml & 1.0 & - & \textbf{7.0} & 0.4 \\
    te & 0.8 & 4.7 & - & \textbf{7.1} \\
    ta & \textbf{8.9} & 4.5 & 0.6 & - \\
    \hline
\end{tabular}
\end{table}

\begin{table}[h]
\centering
  \caption{BLEU score for 4 language pair model 3. The rows are the source language and the columns are the target language. Cells in bold represent the translation directions used in training}
  \label{hi-4langV}
  \begin{tabular}{c|c c c c}
    \hline
       & kn & ml & te & ta \\
    \hline
    kn & - & \textbf{6.5} & 4.5 & 0.8 \\
    ml & 6.8 & - & \textbf{6.4} & 5.5 \\
    te & 0.7 & 2.4 & - & \textbf{6.6} \\
    ta & \textbf{8.1} & 1.7 & 4.5 & - \\
    \hline
\end{tabular}
\end{table}

\subsection{6 language pairs}
2 additional language pairs are added to the 4 unique language pairs of model 2 and a transformer model is trained on all 6 language pairs. The model now sees 1/2 of all possible translation directions during training. Table \ref{hi-6lang} shows the results obtained from the 6 language pair model.\\
We observe that zero-shot translation performance increases drastically. Zero-shot directions now lag by 1.76 BLEU to the trained translation directions.

\subsection{8 language pairs}
Our final model is trained on 8 translation directions, 2 more than the 6 language pair model. The 2 additional language pairs were chosen arbitrarily.\\
Table \ref{hi-8lang} documents the results of the model. We can observe that all translation directions including zero-shot are now above 5 BLEU. Difference in BLEU is inconsistent with zero-shot directions often outperforming the trained translation directions.

\section{Discussion}
\subsection{The Data vs Accuracy Tradeoff}
We observed that as we increased the number of translation directions in the model, the accuracy of the zero-shot translations also increased. Since each language pair has approximately the same number of sentences, the training time increases linearly as we add in more corpora. We found that the convergence time also increased with the addition of more language pairs. The training time per epoch for the 6-language pair model was 50\% higher than the 4-language pair model and the training time per epoch for the 8-language pair model was 100\% higher than the 4-language pair model. In exchange for this increased train time, we noticed that the BLEU score for trained translation directions increased by 0.5-1 points and an increase of approximately 1-3 BLEU for every zero-shot translation direction as we increased the number of language pairs. We noticed that the increase in BLEU becomes minimal as we go past 6-language pairs which could be the optimal threshold for the data vs accuracy tradeoff.

\begin{table}[h]
\centering
  \caption{BLEU score for 6 language pair model. The rows are the source language and the columns are the target language. Cells in bold represent the translation directions used in training}
  \label{hi-6lang}
  \begin{tabular}{c|c c c c}
    \hline
       & kn & ml & te & ta \\
    \hline
    kn & - & \textbf{7.2} & 5.7 & 5.4 \\
    ml & 9.4 & - & \textbf{7.0} & \textbf{7.4} \\
    te & \textbf{10.6} & 6.3 & - & \textbf{7.7} \\
    ta & \textbf{9.2} & 5.4 & 6.3 & - \\
    \hline
\end{tabular}
\end{table}
\subsection{Vanilla Zero-Shot Translation vs Explicit Bridging}
Current research on zero-shot translation recommends using explicit bridging (using a pivot language for zero-shot translation) over vanilla zero-shot translation (single encoder-decoder model). We hypothesize that this recommendation stands only when language relatedness cannot be considered as a supporting factor in multilingual translation. Since our work is limited to Dravidian languages, we can overcome several of the problems that plague zero-shot translations using a shared vocabulary and transliteration and outperform pivot based models.Table \ref{scores} highlights the results obtained and how our models compare against previous baselines. We average the BLEU scores obtained for each target language of translation.\\ 
All our models outperform the previous vanilla zero-shot baseline \cite{kunchukuttan2021empirical} on the WAT-2021 dataset. The only pivot based model that we have to compare our performance against is the multilingual Indic-En and En-Indic model trained by \cite{DBLP:journals/corr/abs-2104-05596}. These models are trained on 11 Indic-En language pairs with each pair having around double the number of sentences that we feed into our models. We utilise the \textit{compute-bleu.sh} script released by \cite{DBLP:journals/corr/abs-2104-05596} to obtained standardized results. 
Our models are unable to beat the BLEU scores of the Samanantar models, but our 6-language and 8-language pair models comes close, with BLEU scores lagging by approximately 3 on average across all language directions despite being a much smaller model that is trained on Indic-Indic language pairs without any implicit bridging. While this does not prove concretely that vanilla zero-shot translations can outperform pivot based method, it does highlight the potential of leveraging linguistic similarity and vocabulary optimizations for zero-shot translations. We believe that with a model of capacity comparable to the Samanantar models \cite{DBLP:journals/corr/abs-2104-05596}, a vanilla zero-shot system can outperform pivot-based models when trained on only a portion of the translation directions.

\begin{table}[h]
\centering
  \caption{BLEU score for 8 language pair model. The rows are the source language and the columns are the target language. Cells in bold represent the translation directions used in training}
  \label{hi-8lang}
  \begin{tabular}{c|c c c c}
    \hline
       & kn & ml & te & ta \\
    \hline
    kn & - & \textbf{7.8} & \textbf{8.4} & \textbf{7.7} \\
    ml & 9.3 & - & \textbf{7.2} & \textbf{7.1} \\
    te & \textbf{10.4} & 6.7 & - & \textbf{7.7} \\
    ta & \textbf{9.1} & 5.8 & 6.9 & - \\
    \hline
\end{tabular}
\end{table}

\begin{table*}[h]
\centering
  \caption{Average BLEU scores when translating into each language for all our models against \textit{Baseline vanilla} zero shot translation\cite{kunchukuttan2021empirical} and \textit{Samanantar pivot}-based baseline \cite{DBLP:journals/corr/abs-2104-05596}}
  \label{scores}
  \begin{tabular}{c|c c c c c c c}
    \hline
      Language & \textit{Baseline Vanilla} & 4-lang-1 & 4-lang 2 & 4-lang 3 & 6-lang & 8-lang & \textit{Samanantar pivot} \\
    \hline
    kn & 0.50 & 5.06 & 3.57 & 5.20 & 9.73 & 9.60 & 13.10 \\
    ml & 0.60 & 5.26 & 5.53 & 3.53 & 6.30 & 6.77 & 10.63 \\
    te & 0.40 & 4.70 & 2.67 & 5.13 & 6.33 & 7.50 & 10.00 \\
    ta & 0.40 & 2.93 & 2.67 & 4.30 & 6.83 & 7.50 & 10.30 \\
    \hline
\end{tabular}
\end{table*}

\FloatBarrier
\begin{table*}[h]
\centering
  \caption{Average BLEU scores when translating into each language for all our models against \textit{Baseline vanilla} zero shot translation\cite{kunchukuttan2021empirical} and \textit{Samanantar pivot}-based baseline \cite{DBLP:journals/corr/abs-2104-05596}}
  \label{scores}
  \begin{tabular}{c|c c c c}
    \hline
      Language & 4-lang & 6-lang & 8-lang & \textit{Samanantar pivot} \\
    \hline
    kn & 5.20 & 9.73 & 9.60 & 13.10 \\
    ml & 3.53 & 6.30 & 6.77 & 10.63 \\
    te & 5.13 & 6.33 & 7.50 & 10.00 \\
    ta & 4.30 & 6.83 & 7.50 & 10.30 \\
    \hline
\end{tabular}
\end{table*}

\subsection{Transliteration and VOLT}
We found that transliteration to an unrelated script (Devanagari) produced more consistency in our results when compared to transliteration to one of the source language scripts (Kannada or Malayalam). Appendix \ref{Appendix} documents the performance of 4 language pair models that were trained after transliteration to a Dravidian language. We observe that the model performs well for the language of the transliteration script. It also performs well for languages that are closely related to the script as is observable in the case of Malayalam performance where we can observe the performance of Tamil to be noticeably better than the other 2 Dravidian languages. We surmise that transliteration to a Dravidian script might introduce a bias in the embeddings of the model. Since Devanagari is an Indo-Aryan script, it results in more consistent and objective model performance.\\
The 4 language pair model trained with a vocabulary generated using VOLT was found to outperform models with a vocabulary of size 32000. This is contrary the popular assumption that Indo-Aryan and Dravidian languages require large vocabulary sizes to perform better in neural machine translation. The limiting of vocabulary might lead to poor performance on out-of-domain corpora, but there is no way to currently test this as we lack standardized test corpora for Dravidian languages. To evaluate our models, we test their performance on the WAT-2021 dataset, which to the best of our knowledge happens to be the only standardized test corpora for the Kannada language.

\section{Conclusions}
This work presents a multilingual Dravidian-Dravidian neural machine translation model for 4 Dravidian languages, that is capable of translating over 12 unique directions with zero-shot learning. We present a simple solution to overcome the shortcomings of zero-shot translation when linguistic similarity can be exploited. We observe the tradeoff between data and zero-shot accuracy and show that a sufficiently good zero-shot translation model can be trained even when we have access to only 2 translation directions. The performance of our single encoder-decoder model is evaluated against state-of-the-art multilingual pivot based approaches. Our model achieves performance on par with previous methods by training on only 4/12 translation directions. Trained directions produce a BLEU score of 6-10 and zero-shot directions BLEU scores that range from poor to on par with trained directions as we increased the number of trained translation directions. We also disprove the theory that Indic languages, and Dravidian languages in particular, require large vocabulary sizes for optimal training, as we show that our VOLT model outperforms the baseline models. We prove that combining transliteration and VOLT can help create data and compute efficient multilingual models that can enable zero-shot translations for morphologically rich languages.

\clearpage

\bibliographystyle{unsrt}  
\bibliography{references}  

\begin{thebibliography}{10}

\bibitem{DBLP:journals/corr/BahdanauCB14}
Dzmitry Bahdanau, Kyunghyun Cho, and Yoshua Bengio.
\newblock Neural machine translation by jointly learning to align and
  translate.
\newblock In Yoshua Bengio and Yann LeCun, editors, {\em 3rd International
  Conference on Learning Representations, {ICLR} 2015, San Diego, CA, USA, May
  7-9, 2015, Conference Track Proceedings}, 2015.

\bibitem{DBLP:journals/corr/WuSCLNMKCGMKSJL16}
Yonghui Wu, Mike Schuster, Zhifeng Chen, Quoc~V. Le, Mohammad Norouzi, Wolfgang
  Macherey, Maxim Krikun, Yuan Cao, Qin Gao, Klaus Macherey, Jeff Klingner,
  Apurva Shah, Melvin Johnson, Xiaobing Liu, Lukasz Kaiser, Stephan Gouws,
  Yoshikiyo Kato, Taku Kudo, Hideto Kazawa, Keith Stevens, George Kurian,
  Nishant Patil, Wei Wang, Cliff Young, Jason Smith, Jason Riesa, Alex Rudnick,
  Oriol Vinyals, Greg Corrado, Macduff Hughes, and Jeffrey Dean.
\newblock Google's neural machine translation system: Bridging the gap between
  human and machine translation.
\newblock {\em CoRR}, abs/1609.08144, 2016.

\bibitem{DBLP:journals/corr/SutskeverVL14}
Ilya Sutskever, Oriol Vinyals, and Quoc~V. Le.
\newblock Sequence to sequence learning with neural networks.
\newblock {\em CoRR}, abs/1409.3215, 2014.

\bibitem{firat-etal-2016-multi}
Orhan Firat, Kyunghyun Cho, and Yoshua Bengio.
\newblock Multi-way, multilingual neural machine translation with a shared
  attention mechanism.
\newblock In {\em Proceedings of the 2016 Conference of the North {A}merican
  Chapter of the Association for Computational Linguistics: Human Language
  Technologies}, pages 866--875, San Diego, California, June 2016. Association
  for Computational Linguistics.

\bibitem{DBLP:journals/corr/JohnsonSLKWCTVW16}
Melvin Johnson, Mike Schuster, Quoc~V. Le, Maxim Krikun, Yonghui Wu, Zhifeng
  Chen, Nikhil Thorat, Fernanda~B. Vi{\'{e}}gas, Martin Wattenberg, Greg
  Corrado, Macduff Hughes, and Jeffrey Dean.
\newblock Google's multilingual neural machine translation system: Enabling
  zero-shot translation.
\newblock {\em CoRR}, abs/1611.04558, 2016.

\bibitem{dong-etal-2015-multi}
Daxiang Dong, Hua Wu, Wei He, Dianhai Yu, and Haifeng Wang.
\newblock Multi-task learning for multiple language translation.
\newblock In {\em Proceedings of the 53rd Annual Meeting of the Association for
  Computational Linguistics and the 7th International Joint Conference on
  Natural Language Processing (Volume 1: Long Papers)}, pages 1723--1732,
  Beijing, China, July 2015. Association for Computational Linguistics.

\bibitem{zoph-knight-2016-multi}
Barret Zoph and Kevin Knight.
\newblock Multi-source neural translation.
\newblock In {\em Proceedings of the 2016 Conference of the North {A}merican
  Chapter of the Association for Computational Linguistics: Human Language
  Technologies}, pages 30--34, San Diego, California, June 2016. Association
  for Computational Linguistics.

\bibitem{zhang-etal-2020-improving}
Biao Zhang, Philip Williams, Ivan Titov, and Rico Sennrich.
\newblock Improving massively multilingual neural machine translation and
  zero-shot translation.
\newblock In {\em Proceedings of the 58th Annual Meeting of the Association for
  Computational Linguistics}, pages 1628--1639, Online, July 2020. Association
  for Computational Linguistics.

\bibitem{DBLP:journals/corr/ChengLYSX16}
Yong Cheng, Yang Liu, Qian Yang, Maosong Sun, and Wei Xu.
\newblock Neural machine translation with pivot languages.
\newblock {\em CoRR}, abs/1611.04928, 2016.

\bibitem{DBLP:journals/corr/ChenLCL17}
Yun Chen, Yang Liu, Yong Cheng, and Victor O.~K. Li.
\newblock A teacher-student framework for zero-resource neural machine
  translation.
\newblock {\em CoRR}, abs/1705.00753, 2017.

\bibitem{DBLP:journals/corr/abs-1811-01389}
Surafel~Melaku Lakew, Quintino~F. Lotito, Matteo Negri, Marco Turchi, and
  Marcello Federico.
\newblock Improving zero-shot translation of low-resource languages.
\newblock {\em CoRR}, abs/1811.01389, 2018.

\bibitem{DBLP:journals/corr/abs-1906-01181}
Jiatao Gu, Yong Wang, Kyunghyun Cho, and Victor O.~K. Li.
\newblock Improved zero-shot neural machine translation via ignoring spurious
  correlations.
\newblock {\em CoRR}, abs/1906.01181, 2019.

\bibitem{DBLP:journals/corr/abs-1903-07091}
Naveen Arivazhagan, Ankur Bapna, Orhan Firat, Roee Aharoni, Melvin Johnson, and
  Wolfgang Macherey.
\newblock The missing ingredient in zero-shot neural machine translation.
\newblock {\em CoRR}, abs/1903.07091, 2019.

\bibitem{DBLP:journals/corr/abs-1805-10338}
Lierni Sestorain, Massimiliano Ciaramita, Christian Buck, and Thomas Hofmann.
\newblock Zero-shot dual machine translation.
\newblock {\em CoRR}, abs/1805.10338, 2018.

\bibitem{DBLP:journals/corr/abs-1907-12437}
Jerin Philip, Vinay~P. Namboodiri, and C.~V. Jawahar.
\newblock A baseline neural machine translation system for indian languages.
\newblock {\em CoRR}, abs/1907.12437, 2019.

\bibitem{10.1145/3407912}
Ashwani Tanwar and Prasenjit Majumder.
\newblock Translating morphologically rich indian languages under zero-resource
  conditions.
\newblock {\em ACM Trans. Asian Low-Resour. Lang. Inf. Process.}, 19(6), oct
  2020.

\bibitem{kunchukuttan2021empirical}
Anoop Kunchukuttan.
\newblock An empirical investigation of multi-bridge multilingual nmt models,
  2021.

\bibitem{DBLP:journals/corr/abs-2104-05596}
Gowtham Ramesh, Sumanth Doddapaneni, Aravinth Bheemaraj, Mayank Jobanputra,
  Raghavan AK, Ajitesh Sharma, Sujit Sahoo, Harshita Diddee, Mahalakshmi J,
  Divyanshu Kakwani, Navneet Kumar, Aswin Pradeep, Kumar Deepak, Vivek
  Raghavan, Anoop Kunchukuttan, Pratyush Kumar, and Mitesh~Shantadevi Khapra.
\newblock Samanantar: The largest publicly available parallel corpora
  collection for 11 indic languages.
\newblock {\em CoRR}, abs/2104.05596, 2021.

\bibitem{xu-etal-2021-vocabulary}
Jingjing Xu, Hao Zhou, Chun Gan, Zaixiang Zheng, and Lei Li.
\newblock Vocabulary learning via optimal transport for neural machine
  translation.
\newblock In {\em Proceedings of the 59th Annual Meeting of the Association for
  Computational Linguistics and the 11th International Joint Conference on
  Natural Language Processing (Volume 1: Long Papers)}, pages 7361--7373,
  Online, August 2021. Association for Computational Linguistics.

\bibitem{NIPS2017_3f5ee243}
Ashish Vaswani, Noam Shazeer, Niki Parmar, Jakob Uszkoreit, Llion Jones,
  Aidan~N Gomez, \L~ukasz Kaiser, and Illia Polosukhin.
\newblock Attention is all you need.
\newblock In I.~Guyon, U.~V. Luxburg, S.~Bengio, H.~Wallach, R.~Fergus,
  S.~Vishwanathan, and R.~Garnett, editors, {\em Advances in Neural Information
  Processing Systems}, volume~30. Curran Associates, Inc., 2017.

\bibitem{DBLP:journals/corr/abs-1904-01038}
Myle Ott, Sergey Edunov, Alexei Baevski, Angela Fan, Sam Gross, Nathan Ng,
  David Grangier, and Michael Auli.
\newblock fairseq: {A} fast, extensible toolkit for sequence modeling.
\newblock {\em CoRR}, abs/1904.01038, 2019.

\bibitem{DBLP:journals/corr/abs-1808-06226}
Taku Kudo and John Richardson.
\newblock Sentencepiece: {A} simple and language independent subword tokenizer
  and detokenizer for neural text processing.
\newblock {\em CoRR}, abs/1808.06226, 2018.

\bibitem{DBLP:journals/corr/SennrichHB15}
Rico Sennrich, Barry Haddow, and Alexandra Birch.
\newblock Neural machine translation of rare words with subword units.
\newblock {\em CoRR}, abs/1508.07909, 2015.

\end{thebibliography}
\clearpage

\appendix
\section{Translation Performance after Dravidian Script Transliteration} \label{Appendix}
This section highlights the performance of our multilingual models when the input data was transliterated to Kannada and Malayalam. All the models here are 4 language pair models trained without VOLT. Table \ref{kn-4lang} shows the performance of Kannada transliteration and Table \ref{ml-4lang} shows the performance of Malayalam transliteration. We observe that despite the Dravidian languages outperforming Devanagari transliteration for some of the language pairs, the performance is inconsistent with a noticeably bad performance for Malayalam-Tamil translation. To prioritise consistency over high BLEU scores, we decided to proceed with Devanagari transliteration for the rest of our experiments.
 \begin{table}[h]
 \centering
  \caption{BLEU score for 4 language pair model with Kannada transliteration. The rows are the source language and the columns are the target language. Cells in bold represent the translation directions used in training}
  \label{kn-4lang}
  \begin{tabular}{c|c c c c}
    \hline
       & kn & ml & te & ta \\
    \hline
    kn & - & \textbf{7.0} & 5.2 & 4.7 \\
    ml & 1.2 & - & \textbf{6.9} & 0.7 \\
    te & 0.4 & 0.6 & - & \textbf{7.0} \\
    ta & \textbf{8.6} & 0.7 & 3.4 & - \\
    \hline
\end{tabular}
\end{table}
\begin{table}[h]
\centering
  \caption{BLEU score for 4 language pair model after Malayalam transliteration. The rows are the source language and the columns are the target language. Cells in bold represent the translation directions used in training}
  \label{ml-4lang}
  \begin{tabular}{c|c c c c}
  \hline
       & kn & ml & te & ta \\
       \hline
    kn & - & \textbf{7.0} & 0.6 & 4.2 \\
    ml & 1.3 & - & \textbf{7.0} & 2.1 \\
    te & 0.5 & 0.8 & - & \textbf{7.3} \\
    ta & \textbf{8.8} & 3.2 & 4.3 & - \\
    \hline
\end{tabular}
\end{table}

\end{document}